\documentclass[letterpaper, 10 pt, conference]{ieeeconf}  

\IEEEoverridecommandlockouts                              
\usepackage[utf8]{inputenc}
\usepackage[T1]{fontenc}
\usepackage{graphicx}
\usepackage{epstopdf}
\usepackage{amsmath}
\usepackage{amssymb}
\usepackage{subfigure}
\usepackage{float}
\usepackage{booktabs}
\usepackage{caption} 
\usepackage{subcaption} 
\usepackage{tabularx}
\makeatletter
\let\NAT@parse\undefined
\makeatother
\usepackage{xcolor}

\definecolor{rblue}{rgb}{0,0.5,1}
\definecolor{hollywoodcerise}{rgb}{0.96, 0.0, 0.63}
\definecolor{lasallegreen}{rgb}{0.03, 0.47, 0.19}
\definecolor{hanpurple}{rgb}{0.32, 0.09, 0.98}
\definecolor{green(pigment)}{rgb}{0.0, 0.65, 0.31}

\usepackage{pifont}
\usepackage{adjustbox}

\usepackage[pagebackref=false,breaklinks=true,colorlinks,bookmarks=false]{hyperref}
\hypersetup{colorlinks,linkcolor={red},citecolor={hanpurple},urlcolor={magenta}}  

\title{\LARGE \bf
Unveiling the Potential of Segment Anything Model 2 for RGB-Thermal Semantic Segmentation with Language Guidance
}

\author{Jiayi Zhao$^{1,2,*}$, Fei Teng$^{1,2,*}$, Kai Luo$^{1,2}$, Guoqiang Zhao$^{1,2}$,  Zhiyong Li$^{1,2}$, Xu Zheng$^{3,\dag}$, and Kailun Yang$^{1,2,\dag}$
\thanks{This work was supported in part by the National Natural Science Foundation of China (Grant No. 62473139), in part by the Hunan Provincial Research and Development Project (Grant No. 2025QK3019), and in part by the Open Research Project of the State Key Laboratory of Industrial Control Technology, China (Grant No. ICT2025B20). 
}
\thanks{$^{1}$The authors are with the School of Artificial Intelligence and Robotics, Hunan University, China.}
\thanks{$^{2}$The authors are also with the National Engineering Research Center of Robot Visual Perception and Control Technology, Hunan University, China.}
\thanks{$^{3}$The author is with the AI Thrust, Hong Kong University of Science and Technology (Guangzhou), China.}%
\thanks{$^{*}$Equal contribution.}
\thanks{$^{\dag}$Corresponding authors: Kailun Yang and Xu Zheng (email: kailun.yang@hnu.edu.cn, zhengxu128@gmail.com).}
}

\begin{document}

\maketitle
\thispagestyle{empty}
\pagestyle{empty}

\begin{abstract}

The perception capability of robotic systems relies on the richness of the dataset. Although Segment Anything Model 2 (SAM2), trained on large datasets, demonstrates strong perception potential in perception tasks, its inherent training paradigm prevents it from being suitable for RGB-T tasks. To address these challenges, we propose SHIFNet, a novel SAM2-driven Hybrid Interaction Paradigm that unlocks the potential of SAM2 with linguistic guidance for efficient RGB-Thermal perception. Our framework consists of two key components: (1) Semantic-Aware Cross-modal Fusion (SACF) module that dynamically balances modality contributions through text-guided affinity learning, overcoming SAM2's inherent RGB bias; (2) Heterogeneous Prompting Decoder (HPD) that enhances global semantic information through a semantic enhancement module and then combined with category embeddings to amplify cross-modal semantic consistency. With $32.27M$ trainable parameters, SHIFNet achieves state-of-the-art segmentation performance on public benchmarks, reaching $89.8\%$ on PST900 and $67.8\%$ on FMB, respectively. The framework facilitates the adaptation of pre-trained large models to RGB-T segmentation tasks, effectively mitigating the high costs associated with data collection while endowing robotic systems with comprehensive perception capabilities. The source code will be made publicly available at \url{https://github.com/iAsakiT3T/SHIFNet}.
\end{abstract}

\section{Introduction} 

The effective and safe operation of autonomous robotic systems relies on accurate pixel-wise classification tasks~\cite{badrinarayanan2017segnet,xie2021segformer,chen2014semantic}, mainly when the precise identification of specific categories directly influences the system’s functionality and decision-making in complex environments. Prior works have achieved accurate perception performance in the RGB modality through sophisticated network designs. Furthermore, to address the limitations of RGB cameras, which make them susceptible to variations in illumination, some researchers~\cite{ha2017mfnet,shivakumar2020pst900} have introduced the thermal modality to compensate for the deficiencies of RGB images in low-light conditions. However, as robotic applications continue to expand across diverse environments, the demand for models with strong generalization capabilities to process heterogeneous and previously unseen data has become increasingly critical. This presents two key challenges for RGB-T perception tasks. 1) Enhancing network performance requires large-scale data collection, which incurs significant labor costs. 2) Fully fine-tuning models on extensive datasets imposes substantial computational demands. Achieving a win-win solution that balances network efficiency and cost-effectiveness remains an urgent problem to be addressed. Fortunately, recent advancements in foundation models have proposed a promising solution to this challenge.

\begin{figure}[!t]
  \centering
  \includegraphics[width=0.48\textwidth]{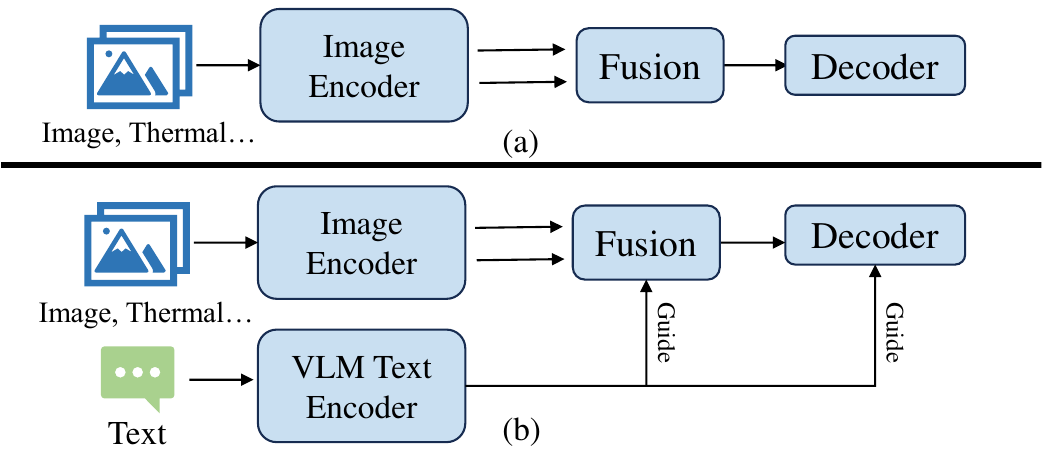}
  \vskip-1ex
  \caption{Architecture comparison: (a) Current framework. (b) our SHIFNet integrating \textbf{pre-trained language-guided RGB-T fusion} and \textbf{semantic hierarchy-aware decoder} to enable knowledge transfer from vision-language models}
  \label{fig:teaser}
  \vskip-4ex
\end{figure}

Segment Anything in Images and Videos (SAM2)~\cite{kirillov2023segment,ravi2024sam}, by leveraging large-scale pre-training on web-scale datasets, demonstrates remarkable generalization across diverse segmentation tasks. If SAM2 can be leveraged for RGB-T perception tasks, it would significantly reduce dependency on task-specific data collection while circumventing computationally prohibitive training procedures~\cite{sam_event,ma2024sam,xie2024pa}. However, research works~\cite{wu2023medical,chen2023sam} indicate that the direct application of SAM2 to multi-modal segmentation yields suboptimal results. This stems from the pre-training paradigm of SAM2, as it is trained on visible images. Directly applying it to multi-modal segmentation tasks overlooks modality-oriented characteristics. Furthermore, despite its strong image encoding capabilities, the fused modalities exhibit feature misalignment, which hinders multi-modal semantic understanding. To overcome these challenges, we introduce a novel framework, as shown in Fig.~\ref{fig:teaser}, termed the SAM2-driven Hybrid Interactive Fusion Paradigm (SHIF).

To fully harness SAM2's segmentation potential for RGB-T tasks, SHIF focuses on two aspects: \textit{Homogeneous feature fusion}, \textit{Heterogeneous Information Matching}. 1) Current feature fusion~\cite{zhang2023delivering,liu2022cmx} methods focus on integrating or selecting~\cite{gemini,wang2022multimodal} features through joint training approaches. They heavily rely on the fine-tuning of network parameters specific to certain datasets, leading to prolonged training durations, constraining the network's ability, and lacking the potential of the SAM2 model. Although some studies~\cite{sam_event,peng2024simple} have introduced adapter-based paradigms based on the SAM series module, they often overlook the inherent modality preferences of SAM pre-trained paradigm, thereby hindering effective information complementation across different modalities. To address this issue, we designed a Semantic-Aware Cross-modal Fusion (SACF) module. SACF leverages the text-image coupling relationships learned by vision-language models~\cite{zhu2023languagebind} through contrastive learning on large-scale datasets and utilizes textual information to dynamically adjust the feature priorities among different visual representations encoded by SAM2. 
It identifies the dominant features and enables dynamically joint visual-textual affinity learning. 2) Although SAM2 exhibits powerful mask generation capabilities, it relies heavily on geometric information (such as shape and boundaries)~\cite{ravi2024sam,opensam}. This bias leads to feature ambiguities in cross-modal semantic understanding and global contextual reasoning, particularly in multi-modal fusion scenarios. To address this issue, we propose a Heterogeneous Prompting Decoder (HPD) with only $3.5M$ parameters. HPD integrates a Semantic Enhancement Module to achieve global semantic information alignment. Then, it leverages category embeddings from a large language model~\cite{zhu2023languagebind} to restructure inter-class relationships, ultimately generating a globally consistent semantic map.

In this paper, we unveil the untapped potential of SAM2 for RGB-Thermal semantic segmentation through language-guided multi-modal adaptation. Extensive evaluations on RGB-T benchmarks (\textit{i.e.}, PST900~\cite{shivakumar2020pst900}, FMB~\cite{liu2023multi}, and MFNet~\cite{ha2017mfnet}) verify our framework's superiority. 
Leveraging SAM2's exceptional segmentation capability, our method ensures comprehensive category-wise segmentation with preserved structural integrity and minimal fragmentation. Even in partial observation scenarios, SAM2's robust foundation maintains complete object delineation, eliminating discontinuities or incomplete contours.

\begin{figure*}[!t]
    \centering
    \includegraphics[width=0.95\linewidth]{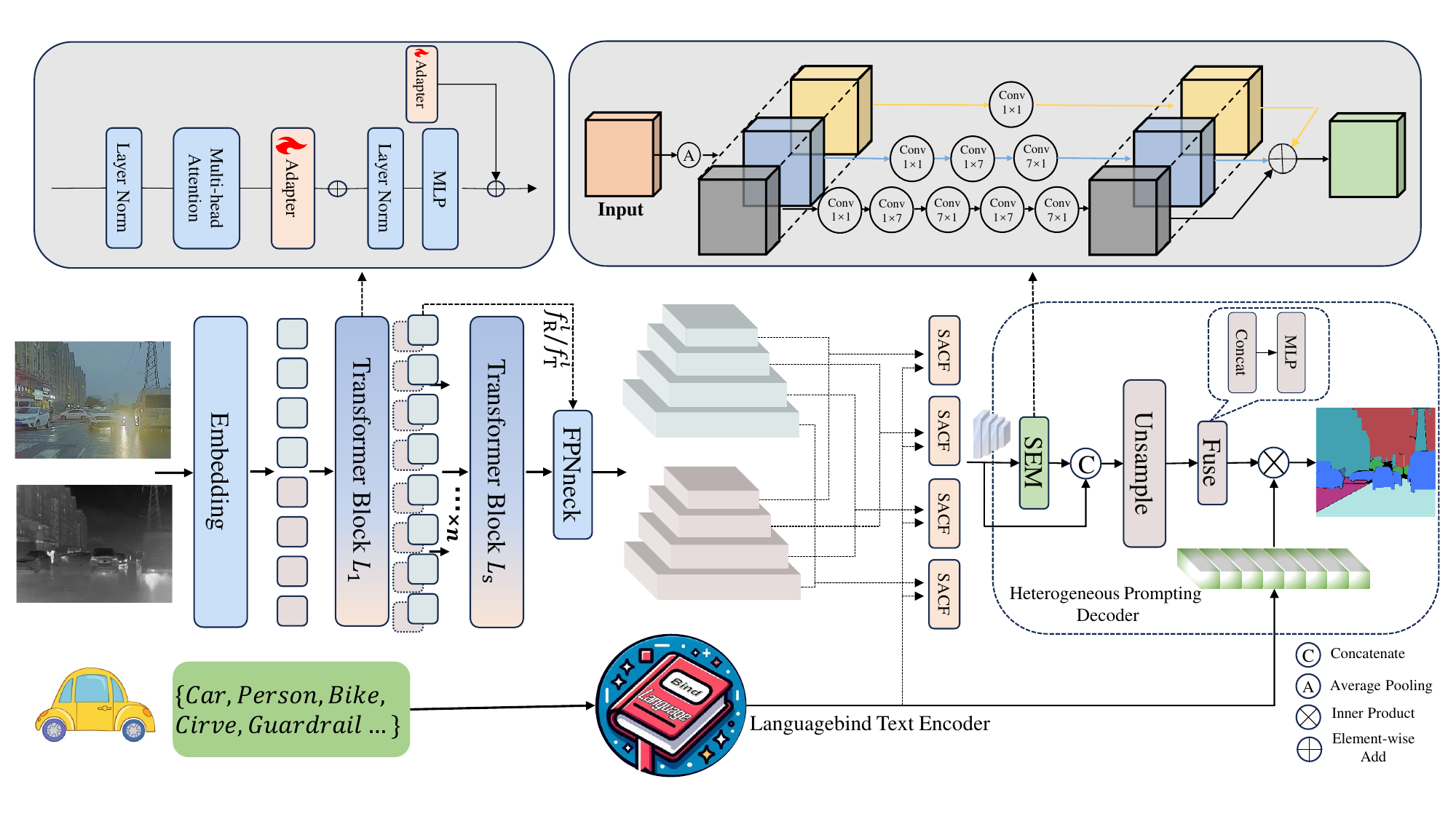}
    \vskip-1ex
    \caption{Architecture of SAM2-driven Hybrid Interactive Fusion (SHIF) paradigm. The framework features: (1) Dual-stream SAM2 encoder with adapters for preserving spectral characteristics, (2) Multi-level Semantic-Aware Cross-modal Fusion (SACF) module, and (3) Heterogeneous Prompting Decoder (HPD) that aligns visual features with semantic embeddings, enabling physics-aware segmentation.}
    \label{fig:Overall-architecture}
    \vskip-3ex
\end{figure*}

The contributions of this work are summarized as follows:
\begin{itemize}
\item[$\bullet$] To meet the perception requirements of intelligent robots across diverse scenarios, we propose the Hybrid Interaction SAM2 paradigm (SHIFNet). SHIFNet unlocks the potential of the SAM2 by addressing modality preferences through language guidance.
\item[$\bullet$] We propose SACF and HPD. SACF dynamically adjusts fusion weights using language embeddings and mitigates modality bias through text-guided feature recalibration. With only $3.5M$ parameters, HPD incorporates global semantic information and utilizes language guidance to enable physics-aware feature reassignment. 
\item[$\bullet$] SHIFNet achieves outstanding performance with $89.8\%$ mIoU on PST900~\cite{shivakumar2020pst900}, $67.8\%$ on FMB~\cite{liu2023multi}, and  $59.2\%$ on MFNet in extreme conditions, demonstrating superior safety-critical perception with $76.5\%$ pedestrian detection accuracy.
\end{itemize}

\section{Related Work}

\subsection{\textit{RGB-T Semantic Segmentation}} 
RGB-T semantic segmentation leverages both RGB and thermal images to improve performance in challenging conditions such as low light or fog~\cite{ha2017mfnet,deng2021feanet}. Early methods relied on simple fusion techniques, like concatenating RGB and thermal features, but struggled with modality misalignment and noise. Recent approaches, including multi-stream networks~\cite{reza2024mmsformer,dong2023egfnet} and multi-scale fusion~\cite{lan2022mmnet,li2024hapnet}, have better captured the complementary nature of RGB and thermal data, enabling more effective feature fusion. Transformer-based models~\cite{reza2024mmsformer}, in particular, have been successful in capturing long-range dependencies and enhancing cross-modal interactions. Integrating foundation models presents a promising win-win solution for RGB-T perception tasks. By leveraging the powerful encoding capabilities of large models, we can mitigate the high cost of data collection in multi-modal perception and reduce the computational burden of network tuning. However, due to their inherent training paradigm, large models exhibit a preference for RGB images.

\subsection{\textit{Segment Anything (SAM) Family}} \label{sec2:1}
The Segment Anything Model (SAM)~\cite{kirillov2023segment} has redefined segmentation through its promptable architecture, trained on $11M$ images and $1.1B$ masks, achieving unprecedented perception capability. While SAM excels in RGB-centric tasks, its inability to process multi-modal inputs (\textit{e.g.}, depth, thermal) limits robotic applications. Recent extensions like SAM2~\cite{ravi2024sam} introduced temporal memory mechanisms for video segmentation, and MM-SAM~\cite{xiao2024segment} enabled multi-modal processing via LoRA-based adaptation.  
Efforts are now increasing to explore SAM2’s potential in broader domains, including medical imaging~\cite{wu2023medical}, 
multi-modal segmentation~\cite{xiao2024segment}, and open-vocabulary dense prediction~\cite{yuan2024open}.
However, despite these advancements, SAM2's potential in the RGB-Thermal (RGB-T) domain remains underexplored. SAM2 and other variants are primarily optimized for visible light images, and as a result, they struggle with the unique challenges presented by the thermal modality, such as differences in feature representations and sensor characteristics. This leaves a significant gap in performance for applications like thermal-aware segmentation, where the fusion of RGB and thermal data is crucial for accuracy, especially in low-visibility environments.
\subsection{\textit{Concept-aware Recognition}} \label{sec:2.3}
Concept-aware networks aim to bridge the gap between visual features and linguistic context, showing great potential in the field of vision-language models. Early works~\cite{tan2019lxmert} leveraged text supervision to fine-tune pre-trained visual encoders for downstream tasks such as VQA~\cite{antol2015vqa} and Image Captioning~\cite{vinyals2015show}. With breakthroughs in foundational models, models like CLIP~\cite{radford2021learning} and Languagebind~\cite{zhu2023languagebind} have become excellent starting points for downstream tasks by pre-training on large-scale noisy image-text pairs.\par
However, the potential of concept-aware networks has not been fully explored in multi-modal semantic segmentation tasks. Existing methods mostly focus on the direct interaction between visual features and text representations~\cite{xie2024sed,shao2024explore}. 
While they have made initial progress, they have not effectively leveraged text concepts to guide fine-grained spatial understanding. In our work, we propose a Heterogeneous Prompting Decoder (HPD) that enables a deep fusion of linguistic context and visual features, addressing the reassignment of fused features encoded by SAM2, unlocking the potential of SAM2 for RGB-T segmentation tasks. 

\section{Methodology}
\subsection{\textit{Overall Architecture}}
The proposed framework, illustrated in Fig.~\ref{fig:Overall-architecture}, is a novel approach that seamlessly integrates multi-modal inputs to achieve robust and efficient semantic segmentation. The core innovation of our approach lies in leveraging linguistic prompts to guide SAM's feature extraction capabilities for multi-modal fusion and driven decoding, unlocking its full representational power for precise RGB-T segmentation. \par
We utilize the SAM2 image encoder for feature extraction across different modalities and share weights. This shared architecture allows the model to leverage the complementary strengths of each modality and learn meaningful representations without the need for separate feature extractors for each modality. The input flow is formally defined by the following:
\begin{equation}
X_{input}=[X_R,X_T].
\end{equation}
Firstly, the RGB and thermal inputs are concatenated and jointly fed into the encoder to obtain the respective features:
\begin{equation}
F_R,F_T={\text{SAM2}}(X_{input}).
\end{equation}
Here, $F_R{=}\{f_R^i\}(i{=}1,2,3,4)$ and $F_T{=}\{f_T^i\}(i{=}1,2,3,4)$ represent the extracted four-level pyramid features from the RGB and thermal modalities, respectively. These multi-level features are then fused through the integration of corresponding pyramid-level features from both modalities:
\begin{equation}
F^i=SACF^i(f_R^i,f_T^i).
\end{equation}
After that, the fused features $F{=}{F^i}(i{=}1,2,3,4)$, along with category embeddings $E_{cls}$ are fed into the decoder for semantic segmentation. This end-to-end flow is designed to extract both low-level spatial information and high-level semantic context. Therefore, the fused features pass through a context module to capture global semantic information, which is then combined with the original fused features to produce enriched representations. These enriched features, together with category embeddings, are then used by the decoder to generate the final segmentation map:
\begin{equation}
Segmap=HPD(F^i,E_{cls}).
\end{equation} 
This architecture utilizes both spatial and semantic features, which are critical for high-quality segmentation and ensures that both visual features and linguistic context are effectively integrated.

\subsection{\textit{Segment Anything Model (SAM2)}}
To ensure the model's scene understanding ability across various scenarios and to avoid the labor-intensive process of multi-modal data collection and calibration, we introduce SAM2 as the feature encoder. Structurally, we extend the original SAM2 image encoder, which was designed to process single RGB inputs, by enabling it to handle multiple modalities through shared weights. Additionally, we fine-tune SAM2's image encoder following the method proposed by Wu \cite{wu2023medical}. This approach allows the model to process diverse inputs while maintaining a unified architecture.\par
SAM2 incorporates Hiera~\cite{ryali2023hiera} as the backbone network and introduces FpnNeck~\cite{lin2017feature} as the Feature Pyramid Network (FPN), including Tiny, Small, Base, and Large. 
In this study, the large scale is utilized.

\subsection{\textit{Semantic-Aware Cross-modal Fusion Module}}
\begin{figure}[!t]
\centering
\includegraphics[scale=0.7]{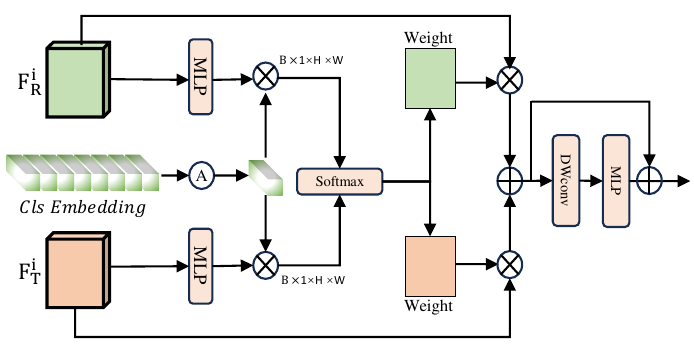}
\vskip-1ex
\caption{Illustration of the SACF module. The hierarchical architecture aligns RGB and thermal features via dual-branch MLP projection. Fused features are progressively refined via gated aggregation and cascaded CXBlocks across four pyramid levels to enhance complementary multi-modal interactions.}
\label{fig:sacf}
\vskip-4ex
\end{figure}

To achieve fine-grained interaction between multi-modal features and semantic priors, we propose a Semantic-Aware Cross-modal Fusion (SACF) module that dynamically adjusts fusion weights through joint visual-textual affinity learning. 
As depicted in Fig.~\ref{fig:sacf}, this hierarchical architecture operates across four pyramid levels, enabling progressive refinement of complementary multi-modal information.\par
Given aligned RGB and thermal features $\{f_R^i,f_T^i\}\in$ $\mathbb{R}^{B\times C\times H\times W}$ at level $i$, we first establish cross-modal semantic correlations through dual-branch projection. Specifically, both modalities undergo nonlinear transformation via parallel MLPs to align their representations with linguistic embeddings:
\begin{equation}
\hat{f}_R = \mathcal{P}{r}(f_R^i) \in \mathbb{R}^{B \times HW \times D},
\end{equation}
\begin{equation}
\hat{f}_T = \mathcal{P}{t}(f_T^i) \in \mathbb{R}^{B \times HW \times D},
\end{equation}
where $\mathcal{P}{r}$ and $\mathcal{P}{t}$ are modality-specific projection layers that map features into a shared semantic space $(D{=}768)$ compatible with language embeddings. After that, the cross-modal attention weights are obtained using condensed text representations. First, we average pool the $N$-class text embeddings $E_{cls}\in \mathbb{R}^{N\times D}$ to obtain global semantic guidance:
\begin{equation}
\bar{E}_{\text{cls}} = \frac{1}{N}\sum_{n=1}^N E_{\text{cls}} \in \mathbb{R}^{1\times D}.
\end{equation}
To address the dynamic variation of modality reliability across spatial regions under changing scenarios (e.g., RGB dominance in well-lit areas versus thermal superiority in low-light conditions), our fusion weights derived from text-guided semantic correlations adaptively allocate modality importance at each position. We compute $\hat{f}_R$ and $\hat{f}_T$ position-wise relevance scores through normalized inner products:
\begin{equation} 
A_R = \frac{\hat{f}_R \cdot \bar{E}{cls}^\top}{|\hat{f}_R| |\bar{E}{cls}|}, \quad A_T = \frac{\hat{f}_T \cdot \bar{E}{cls}^\top}{|\hat{f}_T| |\bar{E}{cls}|}.
\end{equation}
The final fusion weights are obtained by adaptive recalibration:
\begin{equation}
W_R, W_T = {Softmax}({Cat}(A_R, A_T)) \in \mathbb{R}^{B \times 2 \times H \times W}.
\end{equation}
The fused feature $F^i$ at level $i$ is then computed through gated aggregation and progressive refinement:
\begin{equation}
F^i = \Gamma_{\text{out}}\left(\sum_{k=1}^K \mathcal{C}_k\left(\mathcal{G}(W_R \odot f_R^i + W_T \odot f_T^i)\right)\right),
\end{equation}
where $\mathcal{G}(\cdot)$ denotes feature concatenation, $\mathcal{C}_k$ represents a series of CXBlocks~\cite{liu2022convnet}, and $\Gamma_{\text{out}}$is the output projection layer.

\subsection{\textit{Heterogeneous Prompting Decoder}}
To bridge the gap between visual features and linguistic context, the Heterogeneous Prompting Decoder (HPD) is designed to facilitate the guided generation of semantic segmentation maps based on textual descriptions. 
This module integrates category embeddings derived from a language model, leveraging the semantic guidance to adjust feature representations and improve segmentation accuracy. As illustrated in Fig.~\ref{fig:Overall-architecture}, the HPD operates in three key stages:\par
1. Category Embedding. The language model encodes semantic categories into dense vectors $E_{cls}$, where each class label is transformed into a high-dimensional embedding. This embedding serves as a prior for class-wise feature learning in the decoder:
\begin{equation}
E_{cls}=L_{LB}(cls).
\end{equation}
In this context, $L_{LB}$ indicates text encoder from LanguageBind~\cite{zhu2023languagebind}. The obtain class embeddings is $E_{cls}\in \mathbb{R}^{N\times D}$, where $N$ is the number of categories, and $cls$ represents the semantic categories, for example \textit{car}, \textit{person}, \textit{bike}, \textit{etc.}\par
By integrating textual context into the segmentation process, the model focuses on the most relevant features for each class, leading to more precise segmentation boundaries.\par
2. Enhanced stage. 
\begin{figure}[!t]
\centering
\includegraphics[scale=0.5]{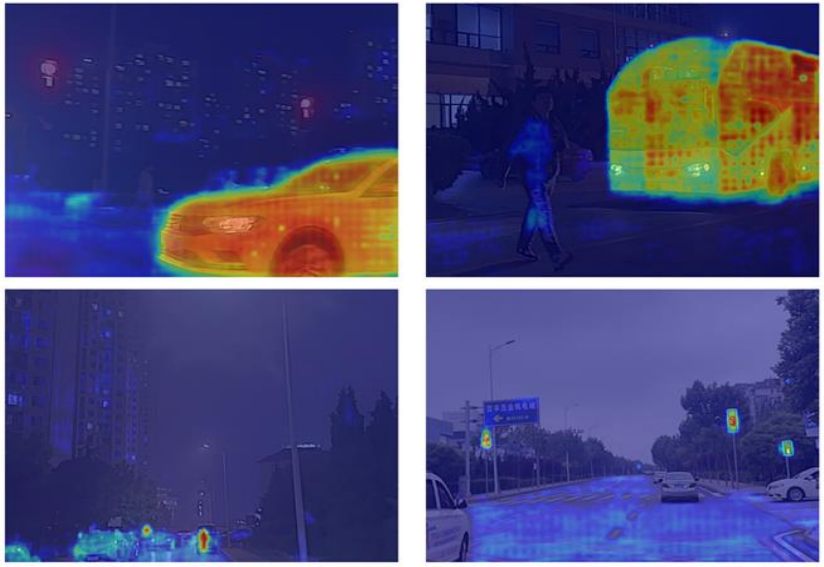}
\vskip-1ex
\caption{Grad-CAM visualizations on the FMB dataset for critical safety categories: Car, Bus, Person, and Traffic Light, highlighting the regions focused on by the model.}
\label{fig:cam}
\vskip-3ex
\end{figure}
To distill the global semantic context, a Semantic Enhancement Module (SEM) is employed as shown in Fig.~\ref{fig:Overall-architecture}. Fused features are first subjected to a global average pooling operation, followed by a $1{\times}1$ convolution. This process stabilizes the feature distributions and suppresses irrelevant background information.
\begin{equation}
F_{avg}=Conv_{1\times1}(Avg(F^i)),
\end{equation}
where $F_{avg}$ represents the features obtained from the enhanced stages. $Conv$ is the abbreviation for the convolutional layer and $AVG$ denotes average pooling. The enhanced features are then processed through three parallel branches, each designed to capture different aspects of the spatial and semantic patterns. These branches focus on lightweight compression, directional modeling, and hierarchical refinement, respectively:
\begin{equation}
F_{b1}=Conv_{1\times1}(F_{avg}),
\end{equation}
\begin{equation}
F_{b2}=Conv_{7\times1}(Conv_{1\times7}(Conv_{1\times1}(F_{avg}))),
\end{equation}
\begin{equation}
F_{b3}=Block_2(Block_1(Conv_{1\times1}(F_{avg})),
\end{equation}
\begin{equation}
Block_i=Conv_{7\times1}(Conv_{1\times7}())(i=1,2)),
\end{equation}
where $\{F_{b1}, F_{b2}, F_{b3}\}$ indicates three different branches. The outputs from different branches are fused via element-wise summation, enhancing salient features while preserving spatial coherence. Two $3{\times}3$ convolutions (dim from $64{\rightarrow}8{\rightarrow}256$) restore high-frequency details and semantic richness, followed by skip-connected concatenation with original pyramid features at corresponding scales and an MLP for dynamic channel recalibration.
\begin{equation}
F_{b} = F_{b1} + F_{b2} + F_{b3},
\end{equation}
\begin{equation}
F_{finally} = Conv_{1\times1}(Up(Concat(F,F_{b}))).
\end{equation}
Hierarchical features are then aligned via bilinear upsampling and concatenated into a $1024$-dimensional representation. 
A parametric $1 {\times} 1$ convolution projects this high-cardinality tensor to $D$, the same as the classification embedding dim.
\par
3. Semantic map. 
Building upon the word-pixel correlation tensor framework (Li~\textit{et al.}~\cite{li2022language}), the final stage involves generating a high-resolution semantic activation map $(F_{finally}\in \mathbb{R}^{\hat{H}\times\hat{W}\times D})$ through joint optimization of visual-textual embeddings.
For each spatial position $(i,j)$, we calculate its similarity to class embeddings via inner product:
\begin{equation}
S_{ij}=F_{finally}^{(i,j)}\cdot E_{cls}^T\in \mathbb{R}^N,
\end{equation}
where $S_{ij}$ represents the semantic affinity scores between the pixel embedding and all classes. This objective maximizes the similarity between pixels and their target class embeddings while suppressing irrelevant class correlations, yielding a high-resolution semantic activation map $A\in \mathbb{R}^{H\times W\times C}$.
\vskip-2ex

\section{Experiments}

\subsection{Datasets}
Our experiments utilize two multi-modal benchmarks with complementary characteristics:
\paragraph{PST900}
PST900~\cite{shivakumar2020pst900} is from the DARPA Subterranean Challenge and consists of $894$ pairs of RGB-T images with a resolution of $1280{\times}720$. The dataset provides annotated segmentation labels for $1$ background class and $4$ foreground classes.

\paragraph{FMB}
FMB~\cite{liu2023multi} captures scenes under a variety of lighting conditions and consists of $1500$ pairs of RGB-T images with a resolution of $800{\times}600$. It provides annotations for $14$ foreground classes and $1$ background class.
\paragraph{MFNet}
MFNet~\cite{ha2017mfnet} is an urban driving scene parsing dataset, containing $1569$ images ($820$ taken at daytime and $749$ taken at nighttime). Eight classes of obstacles commonly encountered during driving are labeled in this dataset.

\paragraph{Metrics and Split}
We use mean Accuracy (mAcc) and mean Intersection over Union (mIoU) as the evaluation metrics for our model. We evaluate $14$ classes for FMB, $5$ classes for PST900 and 9 classes for MFNet, with dataset splits as shown in Table~\ref{tab:dataset}.

\begin{table}[!t]
\caption{\textbf{Overview of dataset splits and image resolutions for PST900, FMB and MFNet.}}
\centering
\setlength{\tabcolsep}{10pt}
\renewcommand{\arraystretch}{1}
\begin{adjustbox}{width=0.48\textwidth}
\begin{tabular}{c|cc|c|c}
\toprule
\textbf{Dataset}  & \textbf{Train} & \textbf{Val} & \textbf{Size} & \textbf{Classes} \\ 
\midrule \hline
PST900~\cite{shivakumar2020pst900}  & 597 & 288 & (1280, 720)& 5 \\ %
FMB~\cite{liu2023multi}   & 1220 & 280 & (1024, 1024) &14 \\ 
MFNet~\cite{ha2017mfnet}   & 1568 & 393 & (640, 480) &9 \\\hline %
\end{tabular}
\end{adjustbox}
\label{tab:dataset}
\vskip-1ex
\end{table}

\begin{table}[!t]
\caption{RGB-Thermal semantic segmentation performance on the PST900 dataset.}
\centering
\label{pst}
\begin{tabularx}{0.4\textwidth}{>{\raggedright\arraybackslash}Xcc} 
\toprule
\textbf{Method} & \textbf{Modalities} & \textbf{mIoU (\%) $\uparrow$} \\
\midrule
UNet~\cite{ronneberger2015u} & RGB-Thermal & 52.8 \\ 
MFNet~\cite{ha2017mfnet} & RGB-Thermal & 57.0 \\ 
PSTNet~\cite{shivakumar2020pst900} & RGB-Thermal & 68.4 \\ 
CACFNet~\cite{zhou2023cacfnet} & RGB-Thermal & 86.6 \\ 
DPLNet~\cite{dong2024efficient} & RGB-Thermal & 86.7 \\ 
MMSFormer~\cite{reza2024mmsformer} & RGB-Thermal & 87.5 \\ 
CRM\_RGBTSeg~\cite{shin2024complementary} & RGB-Thermal & 88.0 \\ 
HAPNet~\cite{li2024hapnet} & RGB-Thermal & 89.0 \\ 
SHIFNet~(Ours) & RGB-Thermal & \textbf{89.8} \\ 
\bottomrule
\end{tabularx}
\vskip-3ex
\end{table}

\subsection{Implementation Details}
All experiments are conducted on two NVIDIA A6000 GPUs using PyTorch 2.1, with mixed-precision training and gradient checkpointing. 
We set a batch size of $4$ ($2$ per GPU) for the FMB, $8$ ($4$ per GPU) for the MFNet and $4$ ($2$ per GPU) for PST900, training for $150$ epochs with the AdamW optimizer ($\beta_1 {=} 0.999$) and a base learning rate of $1 {\times} 10^{-4}$.

\begin{table}[!t]
\caption{RGB-Thermal semantic segmentation performance on the FMB dataset.}
\centering
\label{fmb_result}
\begin{tabularx}{0.4\textwidth}{>{\raggedright\arraybackslash}Xcc} 
\toprule
\textbf{Method} & \textbf{Modalities} & \textbf{mIoU (\%) $\uparrow$} \\
\midrule
GMNet~\cite{michieli2020gmnet} & RGB-Thermal & 49.2 \\
LASNet~\cite{li2022rgb} & RGB-Thermal & 42.5 \\
DIDFuse~\cite{zhao2020didfuse} & RGB-Thermal & 50.6 \\
ReCoNet~\cite{huang2022reconet} & RGB-Thermal & 50.9 \\
U2Fusion~\cite{xu2020u2fusion} & RGB-Thermal & 47.9 \\
SegMiF~\cite{liu2023multi} & RGB-Thermal & 54.8 \\
U3M~\cite{li2024u3m} & RGB-Thermal & 60.8 \\
MMSFormer~\cite{reza2024mmsformer} & RGB-Thermal & 61.7 \\
SHIFNet~(Ours) & RGB-Thermal & \textbf{67.8} \\
\bottomrule
\end{tabularx}
\vskip-2ex
\end{table}

\subsection{Results}
\subsubsection{PST900}
The experimental results on the PST900 underground dataset, as shown in Table~\ref{pst}, demonstrate the exceptional capability of our framework in the safety-critical perception of subterranean environments. 
Achieving a state-of-the-art mIoU of $89.8\%$, our method outperforms HAPNet~\cite{li2024hapnet} by $0.8\%$ absolute improvement, with particularly remarkable gains in life-critical categories. 

\subsubsection{FMB}
The experimental results on the FMB dataset, as summarized in Table~\ref{fmb_result}, demonstrate our framework's exceptional robustness under extreme environmental conditions. Achieving a state-of-the-art mIoU of $67.8\%$, our method outperforms MMSFormer and U3M by $6.1\%$ and $7.0\%$ respectively, with notable improvements in safety-critical categories, as shown in Table~\ref{tab_fwsc}. SHIFNet consistently surpasses existing methods across all classes. This performance gain can be attributed to SHIFNet's effective modality fusion and feature extraction strategies, which enhance accuracy across diverse scenarios.
\subsubsection{MFNet}
Table~\ref{mfnet} presents the RGB-Thermal semantic segmentation performance on the MFNet dataset. The results show that SHIFNet performs well, achieving an mIoU of $59.2\%$, which is slightly lower than CRM\_RGBTSeg and CMXNext, but still outperforms several other methods.
\subsection{Efficiency Analysis}
On the FMB dataset with 480×640 input, our model achieves an average inference time of 123 ms per image on a single NVIDIA RTX A6000 GPU, with a computational cost of 626 GFLOPs.

\begin{table}[!t]
\caption{RGB-Thermal semantic segmentation performance on the MFNet dataset.}
\centering
\label{mfnet}
\begin{tabularx}{0.4\textwidth}{>{\raggedright\arraybackslash}Xcc} 
\toprule
\textbf{Method} & \textbf{Modalities} & \textbf{mIoU (\%) $\uparrow$} \\
\midrule
MFNet~\cite{ha2017mfnet} & RGB-Thermal & 39.7 \\
PSPNet~\cite{zhao2017pyramid} & RGB-Thermal & 46.1 \\
PST900~\cite{shivakumar2020pst900} & RGB-Thermal & 48.4 \\
SegFormer~\cite{xie2021segformer} & RGB-Thermal & 54.8 \\
LASNet~\cite{li2022rgb} & RGB-Thermal & 54.9 \\
SegMiF~\cite{liu2023multi} & RGB-Thermal & 56.1 \\
CMNext\cite{zhang2023delivering} & RGB-Thermal & 59.9 \\
CRM\_RGBTSeg~\cite{shin2024complementary} & RGB-Thermal & 61.4 \\
SHIFNet~(Ours) & RGB-Thermal & 59.2 \\
\bottomrule
\end{tabularx}
\vskip-4ex
\end{table}

\subsection{Ablation Study}
\textbf{Ablation on Key Components of SHIFNet.} 
Table~\ref{Ablation key modules} presents the ablation study of SACF and HPD modules on PST900. Replacing SACF with direct addition and substituting HPD with SegFormer head significantly degrades performance. Method 2 (SACF only) achieves $88.8\%$ mIoU and $93.7\%$ mACC, while Method 3 (HPD only) yields $89.5\%$ mIoU and $93.3\%$ mACC, demonstrating HPD's contribution to segmentation accuracy. Method 4 (SACF + HPD) achieves optimal performance with $89.8\%$ mIoU and $94.1\%$ mACC. The results validate the synergistic effectiveness of SACF in dynamic RGB-T fusion weight adjustment and HPD in physics-aware feature reassignment through language guidance.

\textbf{Ablation on SACF.} 
To validate SACF's efficacy, we systematically ablate its core components against two baseline fusion schemes: 1) Add: Direct element-wise summation without semantic alignment; 2) Concat: Channel-wise concatenation followed by $1{\times}1$ convolution. 
As indicated in Table~\ref{Ablation fusion type}, SACF outperforms Add and Concat by $0.4\%$ and $0.3\%$ mIoU on PST900, respectively, demonstrating its capacity to mitigate modality-specific noise. 
Crucially, removing the category-guided dynamic weighting causes $0.7\%$ mIoU degradation on the FMB dataset, highlighting the necessity of our module.

\begin{table}[!t]
\caption{Ablation of SHIFNet on the PST900 dataset.}
\centering
\label{Ablation key modules}
\begin{tabularx}{0.4\textwidth}{>{\raggedright\arraybackslash}ccccc}
\toprule
\textbf{Method} & \textbf{SACF} & \textbf{HPD} & \textbf{mIoU (\%)}& \textbf{mACC (\%)}\\
\midrule
     1        &  \ding{55} & \ding{55}  & 88.7    & 94.2          \\
     2        &  \ding{51} & \ding{55}  & 88.8    & 93.7          \\ 
     3        &  \ding{55} & \ding{51}  & 89.5    & 93.3        \\ 
     4        &  \ding{51} & \ding{51}  & 89.8    & 94.1         \\ 
\bottomrule
\end{tabularx}
\vskip-2ex
\end{table}

\begin{table}[!t]
\caption{Ablation studies for different fusion methods.}
\vskip-1ex
\centering
\label{Ablation fusion type}
\begin{tabularx}{0.4\textwidth}{>{\raggedright\arraybackslash}cccc}
\toprule
\textbf{Method}& Fusion method & PST900 & FMB\\
\midrule
1 &Add     &  89.4 & 65.5 \\
2 &Concat  &  89.5 & 66.6 \\
3 &SACF w/o text guide    &  88.9 &67.1\\
4 &SACF    &  89.8 &67.8\\
\bottomrule
\end{tabularx}
\vskip-2ex
\end{table}

\begin{table}[!t]
\caption{Ablation studies for different decoders.}
\vskip-1ex
\centering
\label{Ablation decoder}
\begin{tabularx}{0.25\textwidth}{>{\raggedright\arraybackslash}ccc}
\toprule
\textbf{Method} & PST900 & FMB\\
\midrule
All-MLP~\cite{xie2021segformer} &  89.1  &66.1\\
PPM~\cite{zhao2017pyramid}  &85.8  &62.4 \\
HPD~(Ours)                         &  89.8 &67.8\\
\bottomrule
\end{tabularx}
\vskip-2ex
\end{table}

\begin{table}[!t]
\caption{Ablation studies for SEM in HPD.}
\vskip-1ex
\centering
\label{Ablation SEM}
\begin{tabularx}{0.25\textwidth}{>{\centering\arraybackslash}X >{\centering\arraybackslash}X}
\toprule
\textbf{Method} & PST900 \\
\midrule
HPD w/o SEM &  56.6 \\
HPD (Ours)&  89.8\\
\bottomrule
\end{tabularx}
\vskip-5ex
\end{table}

\begin{figure*}[!t]
  \centering
  \includegraphics[width=0.95\linewidth]{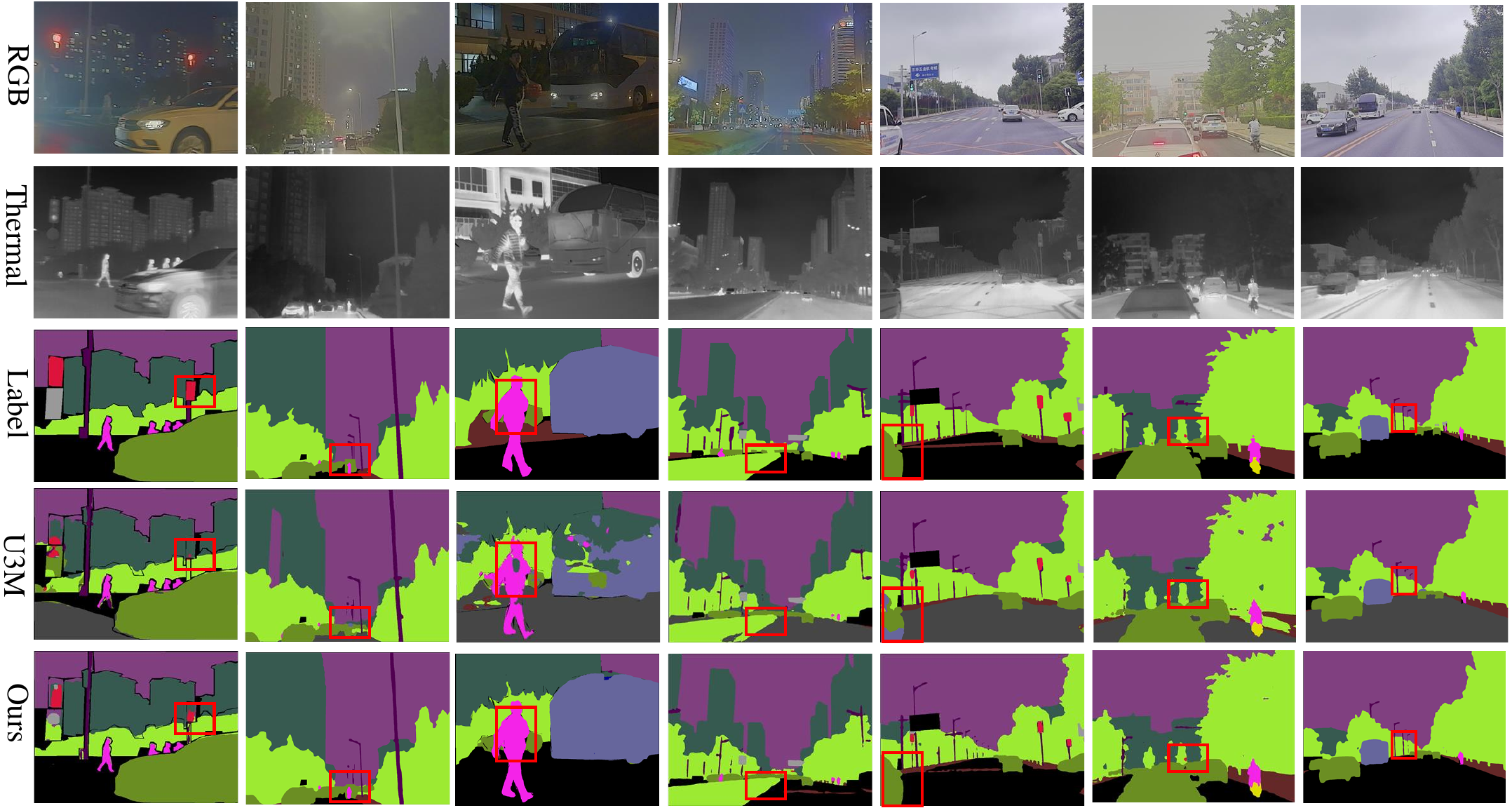}
  \vskip-1ex
  \caption{Day-night segmentation comparison on the FMB dataset: Our method \textit{v.s.} U3M. Critical improvements are shown in low-light conditions with complete pedestrian silhouettes and traffic light recognition while maintaining superior daytime performance through cross-modal RGB-T fusion guided by language priors.}
  \label{fig:vis}
  \vskip-2ex
\end{figure*}

\begin{table*}[!t]
\caption{Per-class semantic segmentation accuracy in mIoU on the FMB dataset~\cite{liu2023multi}.}
\centering 
\label{tab_fwsc}
\begin{tabular*}{0.85\linewidth}{@{}l c c c c c c c c c c@{}}
\toprule
Methods & Modalities & Car & Person & Truck & T-Lamp & T-Sign & Building & Vegetation & Pole & mIoU (\%) \\ 
\midrule
GMNet~\cite{michieli2020gmnet} & RGB-Thermal & 79.3 & 60.1 & 22.2 & 21.6 & 69.0 & 79.1 & 83.8 & 39.8 & 49.2 \\
LASNet~\cite{li2022rgb} & RGB-Thermal & 72.6 & 48.6 & 14.8 & 2.9 & 59.0 & 75.4 & 81.6 & 36.7 & 42.5 \\
DIDFuse~\cite{zhao2020didfuse} & RGB-Thermal & 77.7 & 64.4 & 28.8 & 29.2 & 64.4 & 78.4 & 82.4 & 41.8 & 50.6 \\
ReCoNet~\cite{huang2022reconet} & RGB-Thermal & 75.9 & 65.8 & 14.9 & 34.7 & 66.6 & 79.2 & 81.3 & 44.9 & 50.9 \\
U2Fusion~\cite{xu2020u2fusion} & RGB-Thermal & 76.6 & 61.9 & 14.4 & 28.3 & 68.9 & 78.8 & 82.2 & 42.2 & 47.9 \\
SegMiF~\cite{liu2023multi} & RGB-Thermal & 78.3 & 65.4 & 47.3 & 43.1 & 74.8 & 82.0 & 85.0 & 49.8 & 54.8 \\
U3M~\cite{li2024u3m} & RGB-Thermal & 82.3 & 66.0 & 41.9 & 46.2 & 81.0 & 81.3 & 86.8 & 48.8 & 60.8 \\
MMSFormer~\cite{reza2024mmsformer} & RGB-Thermal & 82.6 & 69.8 & 44.6 & 45.2 & 79.7 & 83.0 & 87.3 & 51.4 & 61.7 \\
SHIFNet~(Ours) & RGB-Thermal & \textbf{86.3} & \textbf{76.5} & \textbf{56.1} & \textbf{47.7} & \textbf{81.0} & \textbf{87.1} & \textbf{88.8} & \textbf{57.1} & \textbf{67.8} \\
\bottomrule
\end{tabular*}
\vskip-3ex
\end{table*}

\textbf{Ablation on HPD.} 
Table~\ref{Ablation decoder} presents experimental results comparing different decoders. 
We performed comparisons with the All-MLP decoder from SegFormer~\cite{xie2021segformer} and PPM from PSPNet~\cite{zhao2017pyramid}. 
The PPM decoder, which generates multi-scale feature maps from the final layer's features rather than utilizing multi-scale features from the backbone, demonstrates poor performance. While SegFormer's All-MLP decoder achieves competitive accuracy ($89.1\%$ in mIoU), it trails our HPD by $0.7\%$ mIoU on PST900 and $1.7\%$ mIoU on the FMB dataset. This highlights the superior performance of HPD in effectively utilizing multi-scale features and enhancing segmentation accuracy. \par
We also validate the effectiveness of the SEM module. 
As shown in Table~\ref{Ablation SEM}, the performance drops significantly on the PST900 dataset when SEM is excluded, underscoring the essential role of SEM in achieving high accuracy.

\subsection{Qualitative Analysis}
Our method excels in robotic perception under challenging conditions such as fog and low-light, ensuring accurate segmentation and object recognition through multi-modal fusion. This robustness is vital for autonomous robots operating in real-world environments, enhancing safety and reliability in dynamic and adverse settings. As depicted in Fig.~\ref{fig:cam} and Fig.~\ref{fig:vis}, the model successfully handles complex scenarios. The integration of SAM2's robust segmentation foundation enables our method to maintain complete object delineation even in partial observation scenarios, eliminating mid-segment discontinuities or incomplete contours across all semantic categories.\par

The prediction of U3M, while showing some degree of accuracy, demonstrates notable limitations, especially in scenarios involving low-light conditions or ambiguous objects like traffic lights and pedestrians. These weaknesses are particularly evident in the misclassification of critical areas under challenging lighting. In contrast, our result shows significant improvements, with our method achieving a more precise focus on critical regions. Even under low-light conditions, our model excels in segmenting these objects accurately, highlighting its superior robustness in handling diverse environmental challenges. This performance enhancement is essential for real-world applications where perception systems must maintain high reliability despite environmental factors such as lighting and noise, positioning our approach at the forefront of autonomous perception technologies.
\vspace{-1mm} 
\section{Conclusion}
In this work, we introduce the SAM2-driven Hybrid Interaction Paradigm (SHIFNet) to empower SAM2 for multi-modal semantic segmentation tasks. SHIP overcomes the challenges posed by manual effort and the limitations of existing datasets. We propose the Semantic-Aware Cross-modal Fusion (SACF) module, which dynamically adjusts the primary modality to mitigate inter-modal bias introduced by SAM2 with language prompt. Additionally, the Heterogeneous Prompting Decoder (HPD) integrates a Semantic Enhancement Module (SEM) to achieve global semantic information alignment and utilizes language guidance to enable physics-aware feature reassignment.
Future work will aim to further enhance SHIFNet's generalization ability to handle diverse real-world scenarios without the need for extensive task-specific data collection, extending its application across various domains requiring robust scene understanding.

\bibliographystyle{IEEEtran}
\bibliography{reference.bib}

\end{document}